\documentclass[letterpaper, 10 pt, conference]{ieeeconf}  

\IEEEoverridecommandlockouts                              
\title{\LARGE \bf
Safe Returning FaSTrack with Robust Control Lyapunov-Value Functions
}

\usepackage{graphicx}
\usepackage{booktabs}
\usepackage{caption}
\usepackage{subfigure}
\usepackage{amsmath} 
\usepackage{amsthm}
\let\labelindent\relax
\usepackage{enumitem}
\usepackage{amsfonts}
\usepackage{amssymb}  
\usepackage{mathtools} 
\usepackage{mathrsfs}
\usepackage{balance}
\usepackage{url}
\usepackage{hyperref}
\usepackage{algorithm}
\usepackage{algpseudocode}
\DeclareMathAlphabet{\mathpzc}{OT1}{pzc}{m}{it}
\usepackage{makecell}

\theoremstyle{definition}

\newtheorem{definition}{Definition}

\newtheorem{theorem}{Theorem}
\newtheorem{remark}{Remark}

\DeclareMathOperator*{\argmin}{arg\,min}

\newcommand{\doi}[1]{\href{http://dx.doi.org/#1}{\normalsize{\textsc{doi:}}~\nolinkurl{#1}}}
\newcommand{\arxiv}[1]{\href{http://arxiv.org/abs/#1}{\normalsize{\textsc{arxiv:}}~\nolinkurl{#1}}}
\usepackage{color}

\usepackage{comment}

\renewcommand{\epsilon}{\varepsilon}

\usepackage[dvipsnames]{xcolor}

\newcommand{\shnote}[1]%
    {\textcolor{magenta}{ #1}}

\newcommand{\goal}{\mathcal{G}}
\newcommand{\pgoal}{\mathcal{G}_p}
\newcommand{\obs}{\mathcal{C}}
\newcommand{\pobs}{\mathcal{C}_p}

\newcommand{\pset}{\mathcal{P}} 
\newcommand{\tset}{\mathcal{X}} 
\newcommand{\tinit}{t}
\newcommand{\tvar}{s}
\newcommand{\thor}{0} 
\newcommand{\tstate}{x} 
\newcommand{\pstate}{p} 

\newcommand{\rstate}{r} 
\newcommand{\rset}{\mathcal{R}}

\newcommand{\ttraj}{\xi_{\tdyn}} 
\newcommand{\rtraj}{\xi}

\newcommand{\tctrl}{u} 
\newcommand{\dstb}{d} 
\newcommand{\pctrl}{u_p} 
\newcommand{\tdyn}{f} 
\newcommand{\pdyn}{h} 
\newcommand{\rdyn}{g} 
\newcommand{\dset}{\mathcal{D}}
\newcommand{\tcset}{\mathcal{U}_s} 
\newcommand{\tcfset}{\mathbb{U}_s} 
\newcommand{\pcset}{\mathcal{U}_p} 
\newcommand{\dfset}{\mathbb{D}}
\newcommand{\ptmat}{Q} 
\newcommand{\lossfunc}{ \ell} 

\newcommand{\valfunc}{V} 

\newcommand{\clvf}{V_\gamma^\infty} 
\newcommand{\minval}{\underline{V}^\infty} 
\newcommand{\TEB}{\mathcal{B}} 
\newcommand{\pTEB}{\mathcal{B}_e} 
\newcommand{\sTEB}{\mathcal{S}} 
\newcommand{\psTEB}{\mathcal{S}_e} 

\newcommand{\rtrans}{\Phi}

\newcommand{\pnext}{p_\text{next}}

\makeatletter
\newcommand\fs@betterruled{%
  \def\@fs@cfont{\bfseries}\let\@fs@capt\floatc@ruled
  \def\@fs@pre{\vspace*{5pt}\hrule height.8pt depth0pt \kern2pt}%
  \def\@fs@post{\kern2pt\hrule\relax}%
  \def\@fs@mid{\kern2pt\hrule\kern2pt}%
  \let\@fs@iftopcapt\iftrue}
\floatstyle{betterruled}
\restylefloat{algorithm}
\makeatother

\newcommand{\zgnote}[1]%
    { \textbf{\textcolor{purple}{#1}} \newline}

\author{\parbox{3 in} \centering  Zheng Gong*, Boyang Li*, Sylvia Herbert
\thanks{ }
}

\author{Zheng Gong$^*$, Boyang Li$^*$ and Sylvia Herbert
\thanks{This research is supported by ONR YIP (\#N00014-22-1-2292) and the UCSD JSOE Early Career Faculty Award. *Both authors contributed equally to this work. All authors are in Mechanical and Aerospace Engineering at UC San Diego 
\{{\href{mailto:zhgong@ucsd.edu}{zhgong},\href{mailto:bol025@ucsd.edu}{bol025},  \href{mailto:sherbert@ucsd.edu}{sherbert}\}@ucsd.edu.}
}%
} 

\usepackage[noadjust]{cite}

\begin{document}
\maketitle

\begin{abstract}
Real-time navigation in a priori unknown environment remains a challenging task, especially when an unexpected (unmodeled) disturbance occurs. In this paper, we propose the framework Safe Returning Fast and Safe Tracking (SR-F) that merges concepts from 1) Robust Control Lyapunov-Value Functions (R-CLVF) \cite{gong2024robust}, and 2) the Fast and Safe Tracking (FaSTrack) framework \cite{chen2020fastrack}. The SR-F computes an R-CLVF offline between a model of the true system and a simplified planning model. Online, a planning algorithm is used to generate a trajectory in the simplified planning space, and the R-CLVF is used to provide a tracking controller that exponentially stabilizes to the planning model. When an unexpected disturbance occurs, the proposed SR-F algorithm provides a means for the true system to recover to the planning model. We take advantage of this mechanism to induce an artificial disturbance by ``jumping'' the planning model in open environments, forcing faster navigation. 
Therefore, this algorithm can both reject unexpected true disturbances and accelerate navigation speed. 
We validate our framework using a 10D quadrotor system and show that SR-F is empirically 20\% faster than the original FaSTrack while maintaining safety. 
\end{abstract}

\section{Introduction}
Safe control for autonomous systems is a challenging task, particularly for dynamic systems navigating through \textit{a priori} unknown environments. For computational efficiency, many algorithms use a simplified (often kinematic) model of the system to generate a path to goals and around obstacles. A more complex model representing the true robot is then used to track this path. Popular path planning algorithms include Dijkstra's \cite{dijkstra2022note}, A$^*$ \cite{hart1968formal}, Rapidly Exploring Random Trees (RRT) \cite{bruce2002real} and heuristic-based methods \cite{stentz1995focussed, likhachev2005anytime}. 
The tracking controller can be generated using, for example, model predictive control (MPC) \cite{Bouffard:EECS-2012-241,borrelli2017predictive,bravo2006robust}, or control Lyapunov functions (CLFs) \cite{hassan2002nonlinear,artstein1983stabilization,sontag1989universal}. For safety, control barrier functions (CBFs) \cite{cbfsurvey} or Hamilton Jacobi (HJ) reachability analysis \cite{evans_hj, bansal2017hamilton,fisac2015reach} can generate safety filters for the controller. 

However, since the planning is done with a simplified model, the path might not be feasible and safe for the actual robot to track. To address this issue, the algorithms in \cite{barbosa2020provably, manjunath2021safe} directly add a CBF and CLF as a constraint in the path planning algorithm. The work in \cite{li2023governor} uses a reference governor control design that moves an equilibrium point that is selected such that the robot can safely stabilize to it. 


\begin{figure}[t]
    \centering
 \includegraphics[width=\columnwidth]{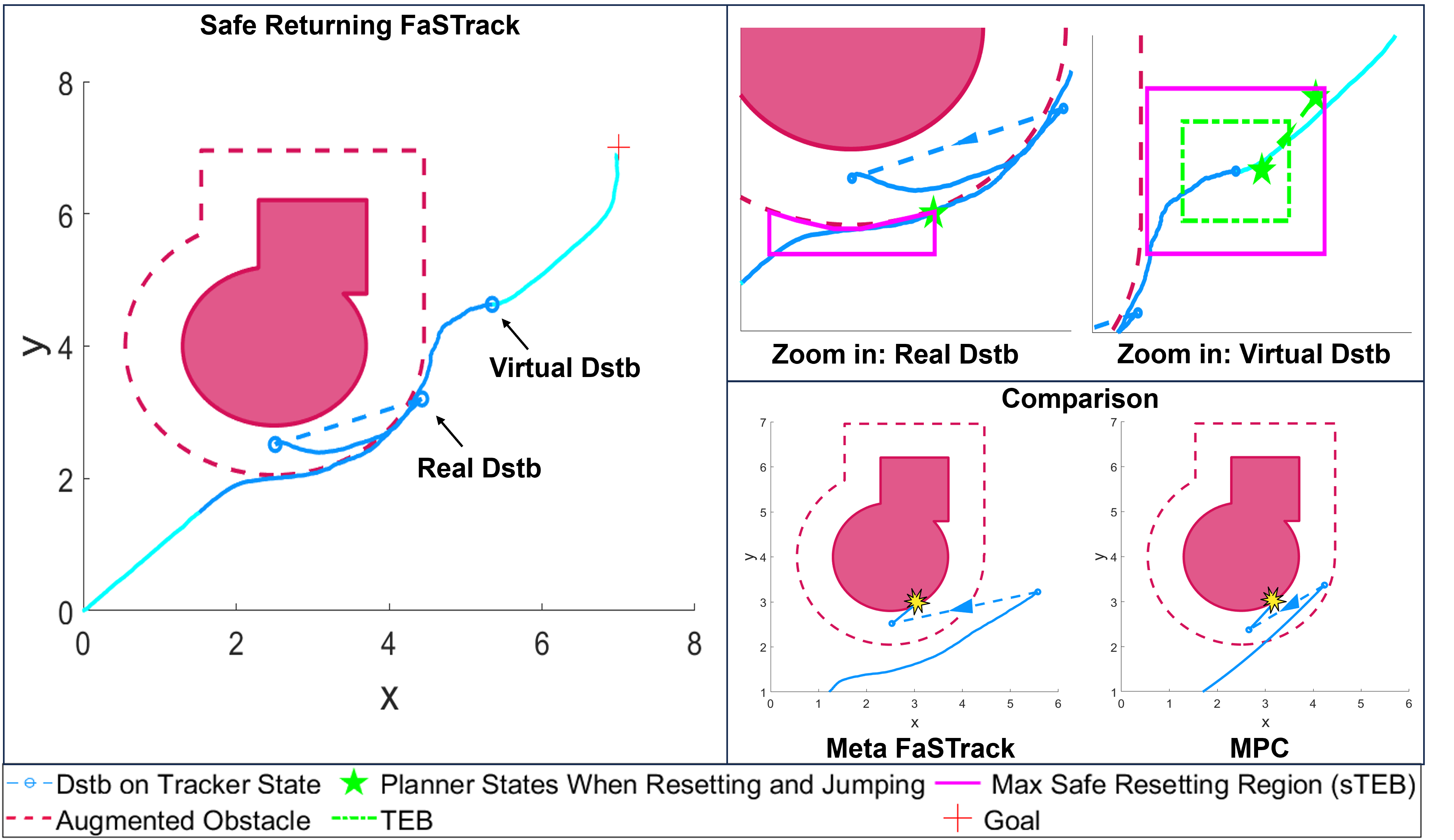}
   \caption{Online simulations for an 8D quadrotor tracking a 2D planning model paired with Rapidly-Exploring Random Trees. Results using SR-F (ours), M-F \cite{fridovich2018planning}, and classic MPC are shown. Top left: the entire trajectory using SR-F. The quadrotor starts at the origin and navigates to a goal (red plus sign). The obstacle (red) is augmented by the tracking error bound. The system's trajectory shown in cyan (fast) and blue (slow). A position disturbance (labeled ``real dstb'') is applied on the quadrotor, pushing it (blue dashed line) close to the obstacle. Top right: zoomed-in views. On the left shows the SR-F algorithm under the real disturbance. The pink region indicates the safe resetting region for the planner (sTEB), which indicates where the planning model may restart to ensure safe convergence. The right shows how the SR-F speeds up navigation in the cyan regions of the trajectory. The algorithm selects the furthest point in the sTEB to reset the planning model, forcing faster navigation while maintaining safety. Bottom right: both M-F and MPC crash after the unexpected disturbance. \vspace{-1em}}

    \label{fig:2D_simulation}
    
\end{figure}

Fast and Safe Tracking (FaSTrack) \cite{ chen2020fastrack} is a modular framework that separates the navigation task into independent planning and tracking tasks (with corresponding planner and tracker models of the autonomous system). Offline, HJ reachability is used to precompute a tracking error bound (TEB) on the maximum deviation that the true tracker model may take from the planner model used for planning. This is paired with an optimal tracking controller that maintains this error bound regardless of the path planning algorithm used by the planner.
Online, the obstacles are augmented with TEB and the path planning algorithm provides a path in the low-dimensional planning space around the augmented obstacles. The tracking controller guarantees the distance between the system and the path is contained in the TEB, and safety is therefore preserved. 

Two main drawbacks of the FaSTrack are 1) the error bound used to augment obstacles is based on worst-case assumptions on the interaction between the planning algorithm and the tracking controller, leading to conservative trajectories, and 2) the framework is unable to deal with unexpected sudden disturbances (i.e. a disturbance larger than the modeled disturbance bound), which may be a common occurrence in uncertain and unstructured environments. PA-FaSTrack \cite{sahraeekhanghah2021pa} and Meta-FaSTrack (M-F) \cite{8460863} mitigate the first drawback, yet no methods are proposed for the second.



We propose the novel Safe Return FaSTrack (SR-F) framework. The main contributions are as follows
\begin{enumerate}
    \item We introduce the SR-F framework, where a CLF-like function in the relative space between the tracker and planner is computed offline, and introduce the new \textit{safe returning} function to accommodate unexpected disturbances. We prove (under specified assumptions) that the SR-F can handle unexpected disturbances and maintain safety. 
    \item We take advantage of this robustness to sudden disturbances by methodically introducing an artificial sudden disturbance by ``jumping'' the planner towards the goal, forcing the autonomous system to speed up in open environments while maintaining safety.
    \item We validate SR-F with a simulated 10D quadrotor navigation task that is subjected to sudden high wind gusts. When no disturbance happens, we show empirically that SR-F speeds up to 20\% compared to the FaSTrack through the jumping process.
\end{enumerate}

\section{Background}



\subsection{Models}
We consider three models: (1) a tracker model that represents the true robot, (2) a planner model that is designed by the user for path planning, and (3) a relative model used to guarantee safety. 
 
\subsubsection{Tracker Model}
The tracker model is given by the following nonlinear ordinary differential equation: 
\begin{align}\label{eq:track_sys}
    \frac{d \tstate}{d \tvar} = \dot \tstate  = \tdyn (\tstate,\tctrl,\dstb), \hspace{3mm} \tstate(\tinit) = \tstate _0, \hspace{3mm} \tvar \in [\tinit,\thor],
\end{align}
where $\tvar$ is the time, $\tstate \in \tset \subseteq \mathbb R^n $ is the tracker state, $\tctrl \in \tcset \subseteq \mathbb R^m$ is the control input, and $\dstb \in \dset \subseteq \mathbb R^d$ is the disturbance. Assume the dynamics $\tdyn: \tset \times \tcset \times \dset \mapsto \tset$ is Lipschitz continuous in $\tstate$ for fixed $\tctrl, \dstb$. Assume the control and disturbance signal $\tctrl (\cdot)$, $\dstb (\cdot)$ are measurable functions: 
\begin{align*}
    &\tctrl(\cdot) \in \tcfset := \{ \tctrl: [\tinit,\thor] \mapsto \tcset, \tctrl (\cdot) \text{ is measurable}\}, \\
    &\dstb (\cdot) \in \dfset := \{ \dstb: [\tinit,\thor] \mapsto \dset, \dstb (\cdot) \text{ is measurable}\},
\end{align*}
where $\tcset$ and $\dset$ are compact sets. Under these assumptions, we can solve for a unique solution of \eqref{eq:track_sys}, denoted as $\ttraj (\tvar ;\tinit,\tstate,\tctrl (\cdot),\dstb (\cdot))$. Denote $\goal \subset \tset$ the goal set, and $\obs \subset \tset$ the constraint set, i.e., the set of states that we want to avoid.

\subsubsection{Planner Model}
The planner model is given by:
\begin{align}\label{eq:plan_sys}
    \frac{d \pstate}{d \tvar} = \dot \pstate = \pdyn(\pstate, \pctrl), \hspace{3mm} \pstate(\tinit) = \pstate_0, 
\end{align}
where $\pstate \in \pset \subseteq \mathbb R^p$ is the planner state, $\pctrl \in \pcset$ is the planner control. Further, assume that $\pset$ is a subspace of $\tset$ and we make analogous assumptions on the planner model dynamics as for \eqref{eq:track_sys} to guarantee a unique solution. 

The goal and constraint sets in the planner space are denoted as $\pgoal \subset \pset$, $\pobs \subset \pset$ respectively. 

\subsubsection{Relative Dynamics}
Define the relative state 
\begin{align} \label{eqn:relative_state}
    \rstate = \rtrans (\tstate,\pstate) (\tstate- \ptmat \pstate),
\end{align}
where $\rstate \in \rset \in \mathbb R^n$, $\ptmat$ augments the planner state and $\rtrans$ is a linear map so that the dynamics can be written as
\begin{align}\label{eq:relative_sys}
    \dot \rstate = \rdyn (\rstate,\tctrl,\pctrl,\dstb).
\end{align}
The existence of $\ptmat$ and $\rtrans$ are justified in \cite{chen2020fastrack}. From the assumption of the tracker and planner model, the relative dynamics also admits unique solution $\rtraj(\tvar ;\tinit,\rstate,\tctrl (\cdot),\pctrl (\cdot),\dstb (\cdot))$. Denote the error states between tracker and planner as $e$ and the rest as $\eta$, i.e., $\rstate = [e,\eta]$.

\subsection{HJ Reachability and Fastrack}
The FaSTrack framework contains two parts: offline computation and online execution. The offline part uses the HJ reachability to generate the TEB, which is a robust control invariant set. Online, it senses the environment, augments the obstacles with the TEB, and then plans and tracks a path around the augmented obstacles.

\subsubsection{HJ reachability (Offline)}

HJ reachability can be formulated and solved as an optimal control problem. Specifically, the cost function $\lossfunc : \rset \mapsto \mathbb R^+$ is designed to measure distance (via the Euclidean norm) in the relative state space. 
The tracker control $\tctrl$ tries its best to track the planner and minimize this cost, whereas the disturbance $\dstb$ and planner control $\pctrl$ try to escape the tracker as far as possible by maximizing this cost. Because the environment and planning algorithm are not necessarily known \textit{a priori}, we assume the worst-case scenario, i.e. that the $\pctrl,\dstb$ can act optimally to $\tctrl$. We define their strategies as mappings $\lambda _p : \tcset \mapsto \pcset$, $\lambda _d : \tcset \mapsto \dset$. We further assume they are restricted to non-anticipative strategies $\lambda _p \in \Lambda_p$, $\lambda _d \in \Lambda_d$ \cite{bansal2017hamilton}. The value function is given by
\begin{align} \label{eqn:value function}
   & \valfunc (\rstate, \tinit) =  \notag \hspace{-1mm} \max _{\lambda _p \in \Lambda _p, \lambda _d \in \Lambda _d} \min _{\tctrl\in \tcfset} \{ \\
   &\max_{s\in [\tinit , \thor]} \ell ( \rtraj( \tvar; \tinit, \rstate, \tctrl (\cdot), \lambda _p(\cdot),\lambda _d(\cdot)) ) \}.
\end{align}
This value function captures the largest cost along one trajectory, with optimal control and disturbance applied. In other words, it captures the worst-case tracking error when the tracker is acting optimally and the disturbance and planner are acting adversarially. We assume the following limit exits on a compact set, and we say the value function converges:
\begin{align} \label{eqn:inf_val}
    \valfunc ^\infty (\rstate) = \lim_{\tinit \rightarrow -\infty} \valfunc (\rstate, \tinit).
\end{align}
The minimal level set of \eqref{eqn:inf_val}, projected into the planner space, is the pTEB. This projection is critical for planning:
\begin{align}\label{eq:TEB}
    \pTEB &:= \{e: \exists \eta \text{ s.t. }\valfunc^\infty(e,\eta) \leq \minval \}. 
\end{align}
Note that if the constraints set $\obs$ is known in advance, we could compute the \textit{inevitable BRT} of the tracker to $\obs$, i.e., the set of states such that the collision must happen~\cite{bansal2017hamilton}. 


\subsubsection{Online Execution} 
The planning module senses the environment, augments the obstacle (if is sensed) with pTEB, and outputs the next planner state. The relative dynamics takes this input and updates the relative state $\rstate$. The tracking module checks its value $\valfunc(\rstate)$ and outputs the tracker control $\tctrl$. This control is sent to the actual robot. The safety is guaranteed \cite{chen2020fastrack} with this online process. 

\begin{remark}
    The value function is computed with a pre-specified disturbance bound $\dset$. A larger $\dset$ corresponds to a larger TEB, which means the obstacles are augmented with a larger set. This causes the augmented environment more dense, which might block all possible paths. However, this also makes the system more robust to the disturbance. On the other hand, a smaller $\dset$ results in a smaller TEB, therefore a more sparse augmented environment, resulting in more flexible choices of paths, but less robust to the disturbance. 
\end{remark}



\subsection{R-CLVF}
Recently, \cite{gong2024robust} proposed the robust control Lyapunov value function (R-CLVF), defined as:

\begin{definition}\textbf{R-CLVF} $\clvf: D_\gamma \mapsto \mathbb R $ of \eqref{eq:relative_sys} is
\begin{align}\label{eq:R-CLVF}
    \clvf (\rstate) = \lim _{\tinit \rightarrow -\infty} \max _{\lambda _p \in \Lambda _p, \lambda _d \in \Lambda _d} \min _{u_s\in \tcfset} \{ \max_{s\in [\tinit , \thor]} e^{\gamma(s-\tinit)}  \ell ( \rtraj (s) ) \}.
\end{align}
Here, $ D_\gamma \subseteq \mathbb R^n$ is the domain of R-CLVF, $\gamma$ is a user-specified parameter which represents the desired decay rate, $\ell (x) = ||x|| - \minval$, and $\minval $ is the minimal value of \eqref{eqn:inf_val}. 
\end{definition}

When $\gamma = 0$, the R-CLVF is just the infinite-time HJ value function \eqref{eqn:inf_val}. Proposition (3) of \cite{gong2024robust} shows that for all $\gamma \geq 0$, the R-CLVFs have the same zero-level set. In other words, for all $\gamma \geq 0$, the zero-level set of the R-CLVFs is the TEB. 

The R-CLVF value of $\rstate$ captures the largest exponentially amplified deviation of a trajectory starting from $\rstate$ to the TEB, under worst-case disturbance. If this value is finite, it means $\rstate$ can be exponentially stabilized to the TEB (Lemma 7 of~\cite{gong2024robust}). 

\begin{theorem} \label{thrm: clvf_exp_decay}
    The relative state can be exponentially stabilized to the TEB from $\mathcal D_\gamma \setminus \TEB$, if the R-CLVF exists in $\mathcal D_\gamma$.
    \begin{align} \label{eqn:exp_decay}
        \min _ {a \in \partial \TEB}||\rtraj(\tvar)-a||  \leq ke^{-\gamma (\tvar-\tinit)} \min _ {a \in \partial \TEB}||\rstate-a||,
    \end{align}
    where $k > 0$ and $\tinit \leq \tvar \leq 0$. 
\end{theorem}

The R-CLVF can be computed by solving the following R-CLVF-VI until convergence
\begin{align*}
     0 = &\max \{ \ell(\rstate) - \clvf(\rstate, t), \notag \\ &\frac{\partial \clvf}{ \partial t} + \min_{\tctrl\in \tcset} \max_{\pctrl \in \pcset,\dstb \in \dset}  \frac{\partial \clvf}{\partial \rstate} \cdot \rdyn(\rstate,\tctrl,\pctrl,\dstb) + \gamma \clvf \}.
\end{align*}
The R-CLVF optimal controller is 
\begin{align} \label{eqn:opt_ctrl}
    \tctrl^* = \argmin_{\tctrl \in \tcset}\max_{\pctrl \in \pcset,\dstb \in \dset} \frac{\partial \clvf}{ \partial \rstate} \cdot \rdyn(\rstate,\tctrl,\pctrl,\dstb).
\end{align}




\section{Safe Returning with Unexpected Disturbance}

\begin{figure} [t]
\centering
\vspace{.5em}
\includegraphics[width=0.4\textwidth]{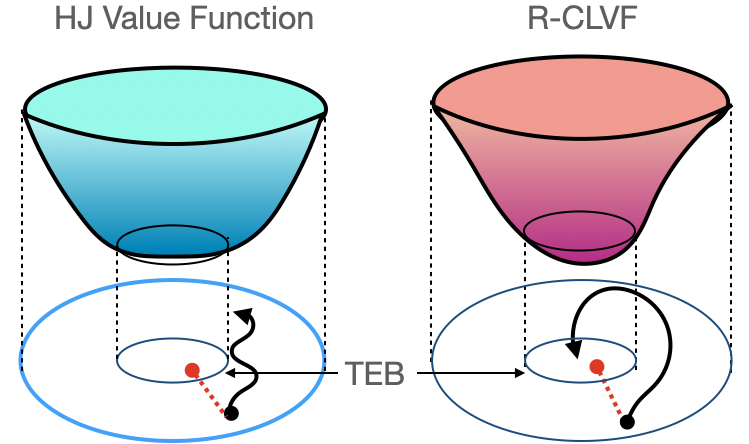}
  \caption{Comparison of relative trajectory using FaSTrack (left) and SR-F (right). The red dotted line denotes an unexpected disturbance that causes the relative state to leave the TEB. The FaSTrack can only guarantee the relative state stays in the current level set, while the SR-F can stabilize the relative state back to the TEB. \vspace{-.5em}}
  \label{fig: fastrack_vs_safe_return_value_functions}
      \vspace{-1em}

\end{figure}

FaSTrack is robust to bounded pre-specified disturbances. However, unexpected and infrequent short-duration disturbances can happen because of communication delays, sudden external forces (e.g. a strong wind), or model mismatch. After a sudden \textit{unexpected} disturbance event that causes the tracker to leave the TEB, the FaSTrack framework only guarantees that the tracker will not exit the \textit{current} level set of the relative value function. This is visualized in Fig.~\ref{fig: fastrack_vs_safe_return_value_functions}, left. The corresponding error bound that must be used to augment obstacles is shown in blue, resulting in conservative plans. 

We propose using the R-CLVF to guarantee that the tracker will not only stay within the current level set but stabilize back to the TEB at the user-specified rate $\gamma$. We present the SR-F framework and highlight two important implications 

\begin{enumerate}
    \item After a sudden unexpected disturbance event, the tracker will converge back to the TEB at an exponential rate $\gamma.$
    \item We can take advantage of this convergence property by introducing an \textit{artificial disturbance} that ``jumps'' the planner forward towards the goal when safe to do so, inducing a faster convergence to the goal.
\end{enumerate}


\begin{figure} 
\begin{center}
\vspace{.5em}
\includegraphics[width=0.4\textwidth]{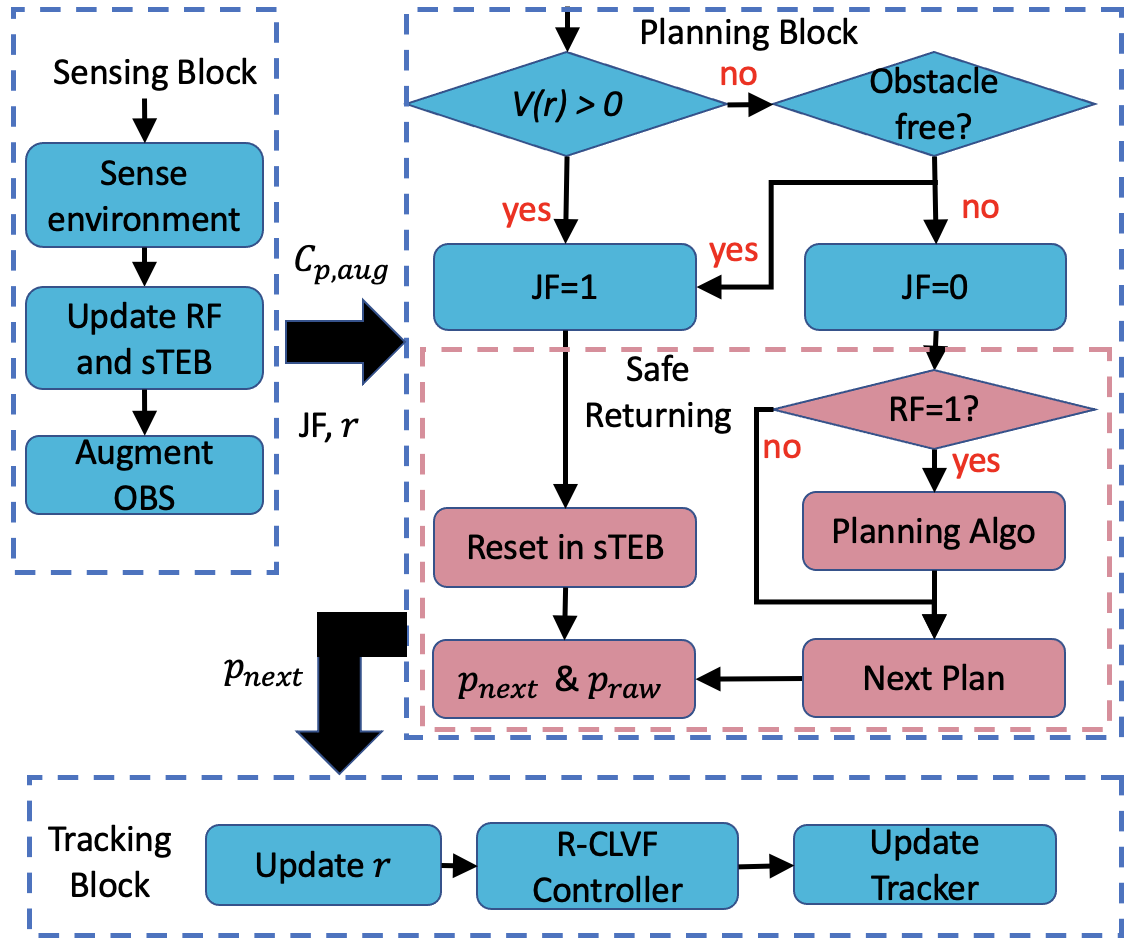}
  \caption{Online flowchart for SR-F. The online algorithm contains three main blocks: the sensing block, the planning block, and the tracking block. The sensing block senses the environment, determines the sTEB and augments the obstacle. The planning block takes in the current tracker state, and does a series of logical judgment. A raw path and next planning state is obtained from the planning block. The tracking block takes in the next plan state, determines the optimal controller, and updates the tracker state.\vspace{-1em}}
  \label{fig: flowchart}
\end{center}
\end{figure}

\subsection{SR-F Algorithm}
The overall algorithm is shown in Alg.~\ref{algo: main}, with a flowchart shown in Fig.~\ref{fig: flowchart}. We begin by explaining this algorithm at a high level. First the ``sensing block'' senses the environment and any unexpected disturbances, then augments obstacles by the \textit {maximum safe resetting region} (sTEB). 

Next the ``planning block'' by default employs a planning algorithm to generate a path through the sensed environment that obeys the dynamics of the planner model. This planning block has modifications for two scenarios: (1) if a sudden disturbance has occurred, the planner may be moved in a way to ensure that the tracker will not hit an obstacle as it converges back to the TEB, (2) if there is an opportunity to do so safely, the planner will ``jump'' ahead towards the goal, forcing the tracker to converge back towards it at the rate $\gamma$.

Finally, there is a ``tracking block,'' which updates the current relative state between the tracker and planner, and applies the pre-computed optimal controller to the tracker that minimizes the distance between itself and the planner.

\subsection{Sensing Block} 


 \textbf{Initialization.} Every iteration starts with checking if the tracker has experienced an unexpected disturbance, which we assume does not cause failure immediately. 
 

\textbf{Environment Sensing.} The robot senses the environment, updates the constraint $\mathcal C_{\text{sensed}}$ (also in the planner space $\mathcal C_{\text{p,sensed}}$), and finds the distance from the tracker to the nearest obstacle within sensing range. This distance is given by  
\begin{align} \label{eqn: distance}
    dst( \tstate ; \mathcal C_{\text{sensed}} ) = \begin{cases}
        \mathcal R & \text{no obstacle} \\
       \min_{a\in \partial \mathcal C_{\text{sensed}}} ||\tstate -a|| & \text{otherwise}
    \end{cases}.
\end{align}
If a new obstacle is sensed, we assign 1 to ReplanFlag (RF).

 \textbf{Computation of Max Safe Resetting Region, sTEB.} Since the planner model is a virtual model with no physical realization, the framework can reset the planner state arbitrarily if needed to ensure that the tracker does not collide with obstacles as it converges back to the planner. We provide a method to find the sTEB, which is denoted as $\sTEB$. Consider a hyperball in the relative state space with radius $dst(\tstate)/2$ and centered at the origin: $B(0,dst(\tstate)/2)$. If the TEB $\TEB$ is contained in this ball $B(0,dst(\tstate)/2)$, the sTEB is largest sub-level set of the R-CLVF contained in $B(0,dst(\tstate)/2)$. Otherwise, the sTEB is the TEB: 
\begin{align} \label{eqn: cTEB}
    \sTEB = 
    \begin{cases}
        \TEB &  \TEB \nsubseteq B(0,dst/2)\\
        \text{largest sub-level set} & \TEB \subseteq B(0,dst/2)
    \end{cases}.
\end{align}
The sTEB in the planner space is given by 
\begin{align}\label{eq:sTEB}
    \psTEB := \{e: \exists \eta \text{ s.t. } [e,\eta] \in \sTEB \}. 
\end{align}

\textbf{Augmentation of Obstacles.} $\psTEB$ is used to augment the obstacles and update the augmented constraint set $\mathcal C_\text{p,aug}$. The outputs of the sensing block are the sensed and augmented obstacle map $\mathcal C_\text{p,sense}$, $\mathcal C_\text{p,aug}$, sTEB, and the RF.

\begin{algorithm} [t]
\caption{SR-FaSTrack}\label{algo: main}
\begin{algorithmic}[1]
\Require  $\clvf$, $\TEB$, sense range $\mathcal R$,  initial states $\tstate_0$, $\pstate_0$. \\
\textbf{Initialization:} 
\State $\tstate \gets \tstate_0$, $\tstate_{\text{old}} \gets \tstate$, $p \gets p_0$, $t \gets 0$, sTEB $\gets \TEB$, JF $\gets 0$, RF $\gets 1$

\While{Goal not reached} 

\State \textbf{Sensing Block} 
\State If unexpected disturbance happens $(\tstate \neq \tstate_{\text{old}})$, update relative state:  
$\rstate \gets  \rtrans (\tstate,\pstate) (\tstate- \ptmat \pstate)$

\State Sense environment, update $\mathcal C_{\text{p, sense}}$, and update distance from the tracker to the obstacle using \eqref{eqn: distance} 
\State RF $\gets 1$ if new obstacle sensed
\State Find safe resetting region $\sTEB$ using \eqref{eqn: cTEB}, augment obstacle with $\psTEB$ and update $\mathcal C_{\text{p, aug}}$ 



\State \textbf{Planning Block}

\If{$\clvf  (r) >0$}  JF $\gets 1$
\ElsIf{ $\clvf (r) \leq 0$ }

\If{ $\mathcal C_{\text{p, sense}}$ is obstacle free} JF $\gets 1$

\ElsIf{Not obstacle free} JF $\gets 0$

 \EndIf 
 \EndIf 
 \State JF, RF, $p_{\text{next}}$, $p_{\text{raw}}$ $\gets$ SafeReturn($\tstate,\pcset$, JF, RF, $p_{\text{raw}}$, $\mathcal C_{\text{p, aug}}$)

 \State \textbf{Tracking Block}
\State $\pstate \gets \pstate_{\text{next}}$, $\rstate \gets \rtrans (\tstate,\pstate) (\tstate- \ptmat \pstate)$ 
 \State $u\gets \argmin_{\tctrl \in \tcset}\max_{\pctrl \in \pcset,\dstb \in \dset} \frac{\partial \clvf}{ \partial \rstate} \cdot \rdyn(\rstate,\tctrl,\pctrl,\dstb)$ 
 \State Update tracker state: $\tstate \gets \text{nextTrack} (\tstate,\tctrl)$
 \State $\rstate,\rstate_{\text{old}} \gets \rtrans (\tstate,\pstate) (\tstate- \ptmat \pstate)$  

\State $ \tvar \gets   \tvar+ \Delta \tvar$
\EndWhile

\end{algorithmic}
\end{algorithm}

\begin{remark}
     To guarantee safety, the consideration of the hyperball $B(0,dst(\tstate)/2)$ is necessary, and its radius must be at least $dst(\tstate)/2$. The reason is that though exponential convergence to the TEB is guaranteed using R-CLVF, it is not necessary that for the next immediate time step, the norm of relative state decreases. This is because of the constant amplifier $k$ in \eqref{eqn:exp_decay}. We illustrate this issue in Fig.~\ref{fig: fastrack_vs_safe_return_value_functions}, right. With the hyperball $B(0,dst(\tstate)/2)$, we guarantee that the distance between the planner and tracker is always smaller than the distance between the planner and the obstacle. 
\end{remark}

\subsection{Planning Block and the Safe Returning Function} 

 \textbf{Jump Evaluation. } The planning block begins by evaluating whether the planner should ``jump'' from its current state. This occurs under two conditions. The first condition occurs when the relative state indicates that the tracker is outside of the TEB (i.e.  $\clvf (r) > 0$).  In this case the planner must jump to ensure that the tracker does not collide with an obstacle as it converges back to the TEB. The second condition is when there are no obstacles within the sensing radius. In this case, the planner creates an artificial disturbance by intentionally ``jumping'' to a further point on its path, increasing the relative state $\rstate$ and forcing the tracker out of the TEB. This induces a speed-up in navigation as the tracker works to converge back at an exponential rate while obeying its control bounds. If either of these conditions for jumping occur, the JumpFlag (JF) is set to 1.
 

 \textbf{Safe Correction Function.} If the JF = 1, the safe correction function sets $\pstate_{\text{next}}$ as the state that is closest to the target, free of the augmented obstacles, and guarantee the relative state is in the sTEB (i.e., $\rtrans (\tstate,\pstate_{\text{next}})(\tstate - Q \pstate_{\text{next}}) \in \sTEB$). We assign 1 to the RF, indicating that the planning algorithm should plan a new path from $\pstate_{\text{next}}$. The JF is reset to 0.

 \textbf{Replan.} If the ReplanFlag has been activated, either from a jump or a new obstacle detected, the path planning algorithm is used to generate a new path for the planner. This path is processed by the function \textit{nextPlan}, which converts the path into a trajectory that obeys the dynamics and control bounds of the planner.  We then reset the RF to 0.

\begin{algorithm}[t]
\caption{Safe Returning Function}\label{algo: safe correction}

\begin{algorithmic}[1]

\Require $\tstate$, $\pcset$, JF, RF, $p_{\text{raw}}$, $\mathcal C_{\text{p,aug}}$ \\
\textbf{Output:} Next plan state $p_{\text{next}}$, $p_{\text{raw}}$, JF, RF



\If{JF = 1}
    \State $p_{\text{next}}$, $p_{\text{raw}}$  $\gets $ the closest point to the target s.t.  $\rtrans (\tstate,\pstate_{\text{next}})(\tstate-Q\pstate) \in $ sTEB and $p \notin \mathcal C_{\text{p,aug}}$
    \State RF $\gets$ 1
\ElsIf{JF = 0}
    \If{ RF = 1}
        \State $p_{\text{raw}} \gets $ PathPlanningAlgo($p,\mathcal C_{\text{p,aug}}$)
    \EndIf
    \State $p_{\text{next}} \gets$ nextPlan($p_{\text{raw}}$, $\pcset$)
    \State remove $ p_{\text{next}} $ from $p_{\text{raw}}$ if $p_{\text{next}} \in p_{\text{raw}} $, otherwise $p_{\text{raw}} \gets p_{\text{raw}}$
    \State RF $\gets$ 0
\EndIf
\State JF $\gets 0$
\State Return $p_{\text{next}}$, $p_{\text{raw}}$, JF, RF
\end{algorithmic}
\end{algorithm}




\subsection{Tracking Block} 
We update the planner state using $p_{\text{next}}$, and update the relative state $\rstate$ using \eqref{eqn:relative_state}. The tracking controller $u$ is determined by \eqref{eqn:opt_ctrl}, which is then sent to the tracker model and updates the tracker state. Note that we keep track of the $\rstate _{\text{old}}$, which is used to check if disturbance happens in the next iteration (lines 21-25 of Algorithm \ref{algo: main}).

\begin{theorem}
    Safety is guaranteed using SR-F if the disturbance does not push the tracker in its \textit{inevitable BRT} of $\obs$. 

    \begin{proof}
        Assume the JF$=$0 for some time step, the SR-F works just like the FaSTrack, and safety is guaranteed \cite{chen2020fastrack}. 

        Assume JF$\neq$0 at some time step. After resetting the planner state and before tracking, denote planner, tracker, and relative states as $\pnext$, $\tstate_1$ and $\rstate_1$. From line 3 of Algorithm \ref{algo: safe correction}, $\pnext$ is chosen such that $\pnext \notin \mathcal C_{\text{p,aug}}$, which means the sTEB centered at $\tstate_1$ is obstacle free. After applying controller \eqref{eqn:opt_ctrl}, denote the new tracker and relative states as $\tstate_2$ and $\rstate_2$. $\rstate_2$ must be contained in a strict subset of the sTEB (by Theorem \ref{thrm: clvf_exp_decay}), which is also obstacle-free. This suggests that $\tstate_2$ is free of obstacles, and safety is guaranteed for the next time step. 

        The overall navigation process is a combination of JF = 1 and JF = 0, and for both cases, immediate safety is guaranteed. We conclude that the whole navigation process is safe concerning modeled and unexpected disturbances. 
    \end{proof}
\end{theorem}

\begin{remark}
    We provide two main benefits compared with the original FaSTrack work. First, SR-F is robust to unexpected disturbances. Second, in the obstacle free region, we mimic a ``beneficial disturbance'' and intentionally make the planner jump to accelerate the whole navigation process. 
\end{remark}

\section{Experiments}

We demonstrate that SR-F can provide safety guarantees given unexpected disturbances, and accelerate the navigation process. We consider two numerical examples: a 10D and an 8D near-hover quadrotor \cite{Bouffard:EECS-2012-241} tracking a 3D and a 2D integrator planner model with a RRT planner \cite{RRT_matlab}, respectively. All simulations are conducted in MATLAB.
\begin{subsection}{Offline computation} 
$\mathit{1) 10D-3D}$: The system dynamics of the 10D quadrotor (tracker) and the 3D integrator (planner) are from Example B~\cite{chen2020fastrack}. The tracker states $(x,y,z)$ denote the position, $(v_x, v_y, v_z)$ denote the velocity, $(\theta_x, \theta_y)$ denote the pitch and roll, $(\omega_x, \omega_y)$ denote the pitch and roll rates. The tracker has controls $(u_x, u_y, u_z)$, representing the desired pitch and roll angle and the vertical thrust. The planner has controls $(\hat{v}_x, \hat{v}_y, \hat{v}_z)$, representing the velocity in each positional dimension. The system parameters are set to be $d_0 = 10, d_1 = 8, n_0 = 10, k_T = 0.91, g = 9.81$, $|u_x|, |u_y| \leq \pi/9$, $u_z\in [0, 1.5g]$, $|\hat{v}_x|,| \hat{v}_y|,| \hat{v}_z | \leq 0.5$, $d_x=d_y=d_z=0$. 


The relative dynamics can be obtained as
\begin{align}\label{eq: relative system}
    &\dot{x}_r = v_x - \hat{v}_x + d_x, \hspace{2mm} \dot{v}_x =  g\tan{\theta_x}, \hspace{2mm} \dot{\theta}_x = -d_1\theta_x + \omega_x, \notag \\ 
    &\dot{\omega}_x =  -d_0\theta_x + n_0u_x, \hspace{2mm} \dot{y}_r =  v_y - \hat{v}_y + d_y, \hspace{2mm}  \dot{v}_y = g\tan{\theta_y}, \notag \\
    & \dot{\theta}_y = -d_1\theta_y + \omega_y, \hspace{2mm} \dot{\omega}_y = -d_0\theta_y + n_0u_y, \notag \\
    & \dot{z}_r = v_z- \hat{v}_z + d_z, \hspace{2mm} \dot{v}_z = k_Tu_z - g.
\end{align}

This is decomposed into three independent subsystems $(x_r, v_x, \theta_x, \omega_x)$, $(y_r, v_y, \theta_y, \omega_y)$, $(z_r, v_z)$ \cite{zheng2023decomp}, allowing us to solve for the R-CLVF more tractably.

$\mathit{2) 8D-2D}$: The relative dynamics of the 8D tracker and the 2D planner are the $x, y$ subsystems above. 
\end{subsection}

\begin{subsection}{Online Planning and Navigation} 
\begin{figure}
    \centering
    \vspace{.5em}
    \includegraphics[width=0.4\textwidth]{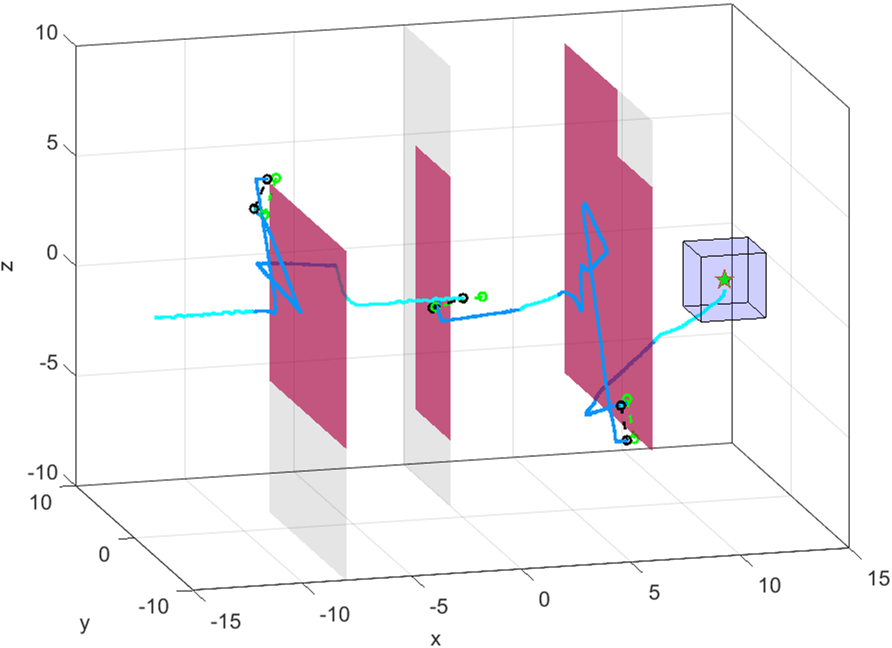}
    \caption{10D-3D simulation using SR-F. The tracker's trajectory switches between cyan and blue, indicating that the tracker jumps (cyan) when obstacle-free, and tracks a RRT path when not obstacle-free. The planner's position is the green star in the translucent blue box (representing sTEB). Both systems start on the left and navigate to a goal on the right.
    The three light grey rectangles are obstacles, and once sensed by the quadrotor they turn red. When the quadrotor is passing near an obstacle, it experiences an unexpected disturbance to its position (black dashed line), mimicking a sudden wind gust. 
    The green dashed line shows the change of the planner's position after replanning in Alg.~\ref{algo: safe correction}.\vspace{-1em}}
    \label{fig:simulation}
        \vspace{-.5em}

\end{figure}



The online simulations for the 8D-2D and 10D-3D are shown in Figs.~\ref{fig:2D_simulation} and  \ref{fig:simulation} respectively. To demonstrate the advantage of the SR-F framework, we compare FaSTrack and M-F with SR-F. We design three experiments with different disturbance settings: $1)$ no disturbance, $2)$ unexpected disturbance to the position states (like a sudden wind), $3)$ unexpected disturbance to the position and velocity states that act in the worst-case. The results are summarized in Table \ref{tab:comparison}.
When no disturbance exists, safety is guaranteed for all three frameworks, and they all reach the goal if no collision happens. When unexpected disturbances exist, both the FaSTrack and M-F collide for more than $80\%$ of runs. 

We highlight that SR-F guarantees safety under unexpected disturbances, though it takes more time to reach the goal. This is because FaSTrack and M-F do not consider the unexpected disturbance, and do not spend time to replan. However, it is preferable to sacrifice the navigation speed for the safety guarantee in most real-world applications. Also, note that the navigation speed is affected by the planning algorithm used and the environment. 


Note that in all simulations, positional disturbances are intentionally given such that the trackers are prone to crash when using M-F and FaSTrack, to demonstrate the safety provided by SR-F even under their failures. When unexpected disturbances are generated by uniformly distributed noise, M-F and FaSTrack have collision rates under $10\%$.
\end{subsection}


\begin{table}[b]
\centering
\caption{Comparison of FaSTrack, M-F and SR-F frameworks for 10D-3D system. Each row averaged across 40 runs.}
\label{tab:comparison}
\resizebox{\columnwidth}{!}{%
\begin{tabular}{@{}c|clc|clc|clc@{}}
\toprule
Types of Disturbance    & \multicolumn{3}{c|}{No Dist}                                                    & \multicolumn{3}{c|}{Pos Dist}                                                 & \multicolumn{3}{c}{Pos + Vel Dist}                                            \\ \midrule
Metrics                 & \multicolumn{1}{c|}{FaSTrack}   & \multicolumn{1}{l|}{M-F}  & SR-F                       & \multicolumn{1}{c|}{FaSTrack}   & \multicolumn{1}{l|}{M-F} & SR-F                      & \multicolumn{1}{c|}{FaSTrack}   & \multicolumn{1}{l|}{M-F} & SR-F                      \\ \midrule
Reach Goal (\%)         & \multicolumn{1}{c|}{100} & \multicolumn{1}{l|}{100} & 100                       & \multicolumn{1}{c|}{15}  & \multicolumn{1}{l|}{12} & 100                      & \multicolumn{1}{c|}{11}  & \multicolumn{1}{l|}{10} & 100                      \\ \cmidrule(r){1-1}
Obstacle Collision (\%) & \multicolumn{1}{c|}{0}   & \multicolumn{1}{l|}{0}   & {\color[HTML]{9A0000} 0}  & \multicolumn{1}{c|}{85}  & \multicolumn{1}{l|}{88} & {\color[HTML]{9A0000} 0} & \multicolumn{1}{c|}{89}  & \multicolumn{1}{l|}{90} & {\color[HTML]{9A0000} 0} \\ \cmidrule(r){1-1}
Navigation Time (s)     & \multicolumn{1}{c|}{96}  & \multicolumn{1}{l|}{77}  & {\color[HTML]{9A0000} 81} & \multicolumn{1}{c|}{102} & \multicolumn{1}{l|}{70} & 115                      & \multicolumn{1}{c|}{101} & \multicolumn{1}{l|}{73} & 121                      \\ \bottomrule
\end{tabular}%
}
\end{table}

\bibliographystyle{IEEEtran}\begin{scriptsize}
\bibliography{ref}\end{scriptsize}

\end{document}